\documentclass[letterpaper, 10 pt, conference, hidelinks]{ieeeconf}
\IEEEoverridecommandlockouts                              % This command is only needed if 
\usepackage{graphicx,dblfloatfix}
\usepackage{amsmath} % assumes amsmath package installed
\usepackage{amssymb}  % assumes amsmath package installed
\usepackage{subcaption}
\usepackage{array}
\usepackage{cite}   % collect references numbers inside single set of square brackets
\usepackage{hyperref}
\usepackage{float}
\usepackage{mathtools}
\usepackage{multirow}
\usepackage{siunitx}

\usepackage{gensymb}

\usepackage{rotating,booktabs}

\usepackage{microtype}

\hypersetup{
    colorlinks=false,
    linkcolor=black,
    urlcolor=black,
}
\def\showcomments{1}
\usepackage{xcolor}
\newcommand{\peter}[1]{{
    \if\showcomments1
        \color{red}PIC: #1
    \fi
}}

\newcommand{\juxicomment}[1]{{
    \if\showcomments1
        \color{red}JL: #1
    \fi
}}
\title{\LARGE \bf Enabling Failure Recovery for On-The-Move Mobile Manipulation}

\author{Ben Burgess-Limerick$^{1}$, Chris Lehnert$^{1}$,  J\"urgen Leitner$^{2}$, Peter Corke$^{1}$% <-this % stops a space
\thanks{This research was supported by the QUT Centre for Robotics.}
\thanks{$^{1}$Ben Burgess-Limerick, Chris Lehnert, and Peter Corke are with the Queensland University of Technology Centre for Robotics (QCR), Brisbane, Australia
        {\tt\small ben.burgesslimerick@qut.edu.au}}
\thanks{$^{2}$J\"urgen Leitner is with LYRO Robotics, Brisbane, Australia}%
}
\begin{document}
\maketitle
\thispagestyle{empty}
\pagestyle{empty}
%%%%%%%%%%%%%%%%%%%%%%%%%%%%%%%%%%%%%%%%%%%%%%%%%%%%%%%%%%%%%%%%%%%%%%%%%%%%%%%%
\begin{abstract}
We present a robot base placement and control method that enables a mobile manipulator to gracefully recover from manipulation failures while performing tasks on-the-move. A mobile manipulator in motion has a limited window to complete a task, unlike when stationary where it can make repeated attempts until successful. Existing approaches to manipulation on-the-move are typically based on open-loop execution of planned trajectories which does not allow the base controller to react to manipulation failures, slowing down or stopping as required. To overcome this limitation, we present a reactive base control method that repeatedly evaluates the best base placement given the robot's current state, the immediate manipulation task, as well as the next part of a multi-step task. The result is a system that retains the reliability of traditional mobile manipulation approaches where the base comes to a stop, but leverages the performance gains available by performing manipulation on-the-move. The controller keeps the base in range of the target for as long as required to recover from manipulation failures while making as much progress as possible toward the next objective. 
See \href{https://benburgesslimerick.github.io/MotM-FailureRecovery}{\url{benburgesslimerick.github.io/MotM-FailureRecovery}} for videos of experiments. 

\end{abstract}

% \begin{keywords}
% Mobile Manipulation, Control Architectures and Planning, Reactive and Sensor-Based Planning.
% % Perception for Grasping and Manipulation, Grasping.
% \end{keywords}

\setcounter{footnote}{2}
\section{Introduction}

Performing multi-step mobile manipulation tasks, such as pick-and-place tasks, with the mobile base in motion has been shown to be an effective means of reducing execution time \cite{ThakarTimeOptimal}. However, performing tasks on-the-move introduces a new challenge for recovering from failures. In particular, if the manipulation is not completed while the robot transits past the target, the robot will be unable to recover without incurring a significant time cost as it returns to the target. 

Existing approaches to manipulation on-the-move are typically based on open-loop execution of a planned trajectory, where a single attempt at the task is planned \cite{ThakarTimeOptimal, ThakarManipulatorMotionPlanning, ThakarUncertainty, Colombo, Zimmermann, XuOptimizationMotionPlanner}. If any failure occurs, such as a poorly executed grasp, a new trajectory is required, which is likely to require significant planning time. 

A close analogue to this problem is grasping an object from a moving conveyor, where there is a limited window for completing the grasp. Existing approaches to conveyor grasping typically do not have time to recover from a failed grasp \cite{Navarro, Islam, Akinola}. A mobile manipulator introduces the possibility to reactively adjust the base motion to ensure that sufficient time is available to complete a manipulation task, while still minimising execution time in multi-step tasks. 

A recent reactive approach \cite{MotM} to manipulation on-the-move has been presented which allows for online repeat attempts in the event of a manipulation failure. In \cite{MotM}, a simple base placement and control system is implemented which does not consider how the base should be best controlled to allow sufficient time for the manipulator to recover from the failure, while also completing the task as quickly as possible.

\begin{figure}[t]
\centerline{\includegraphics[width=0.99\linewidth]{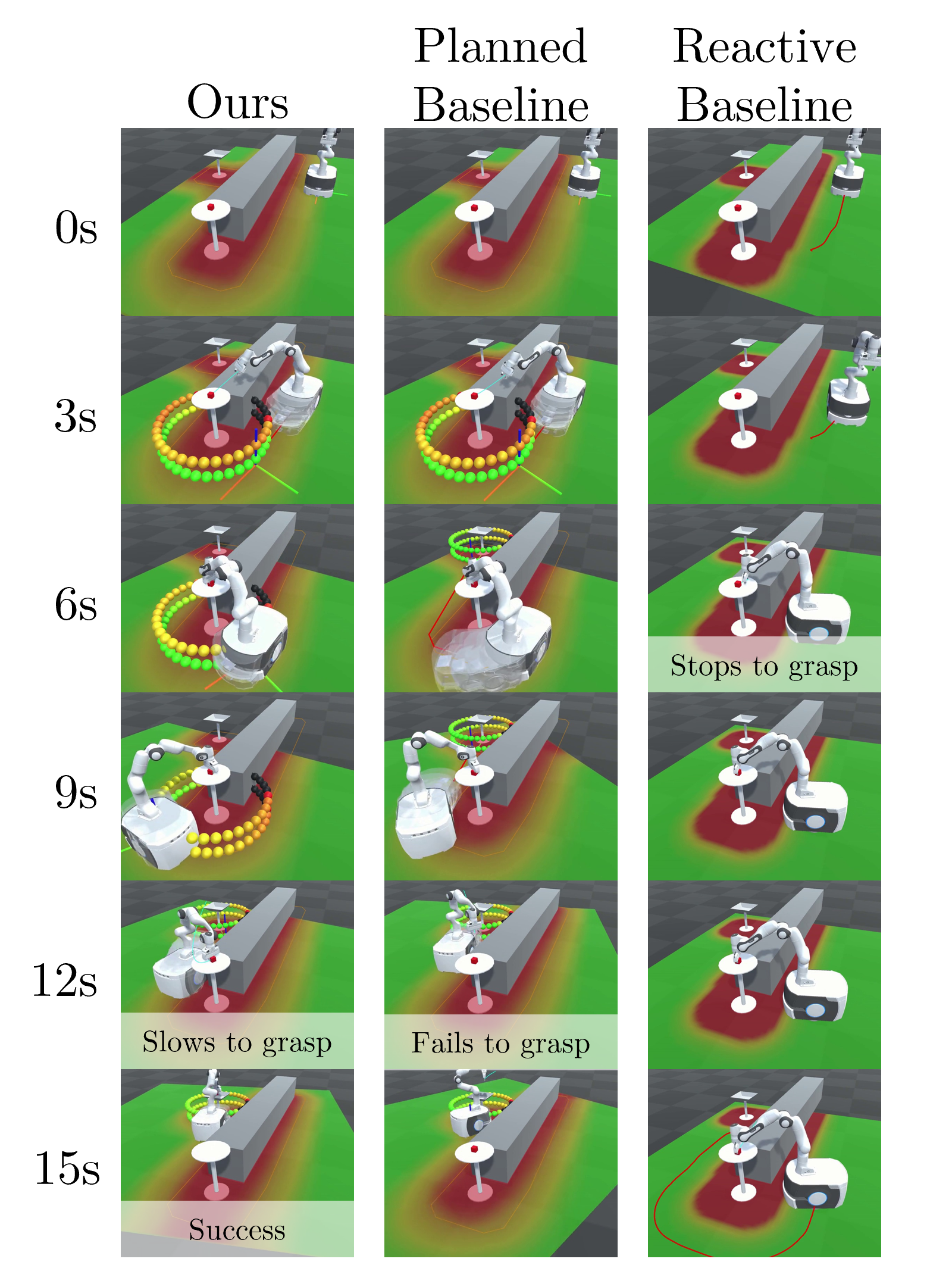}}
\caption{Frames from video of each method performing a pick-and-place task with a grasp failure delay of 6 \si{s}.}
\label{fig:VideoFrames}
\vspace{-20pt}
\end{figure}

In this work we present a reactive base control system that integrates with the architecture presented in \cite{MotM} and allows for failure recovery during manipulations performed on-the-move, while also minimising task execution time. The method is investigated in simulated pick-and-place experiments and compared with two baseline approaches representing existing reactive approaches to mobile manipulation and more recent on-the-move methods (Fig. \ref{fig:VideoFrames}). 

Our results demonstrate the effectiveness of the proposed approach in handling failures and reducing execution time in multi-step mobile manipulation tasks. The proposed method outperforms traditional approaches to mobile manipulation and more recent on-the-move methods in terms of task completion rate and execution time.

\section{Problem Definition}
Graceful recovery from a failure during on-the-move tasks introduces new challenges for mobile manipulator base control. In this work, we focus on pick-and-place tasks where a single object must be grasped and then dropped in a new location. To successfully recover from a failed grasp, the robot must stay within reach of the object until the grasp can be completed. However, there is often an opportunity to improve overall task time by continuing base motion toward the place location for as long as possible. For tasks where multiple objects need to be transported or there are multiple acceptable drop points, the optimal decision may be to change targets entirely. We do not consider these cases in this work. 

\begin{figure}[t]
\begin{subfigure}{.303\linewidth}
\centering
\includegraphics[width=\linewidth]{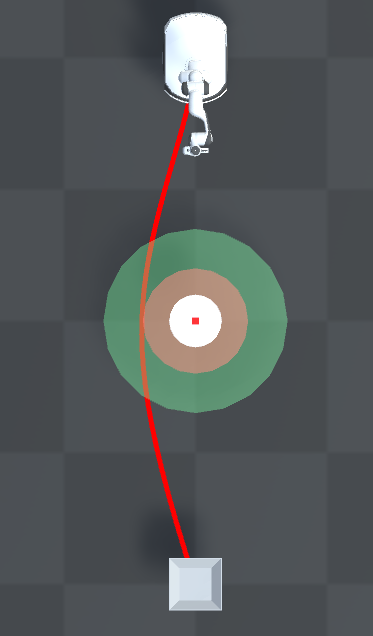}
\caption{Line}
\label{fig:Line}
\end{subfigure}
\begin{subfigure}{.26\linewidth}
\centering
\includegraphics[width=\linewidth]{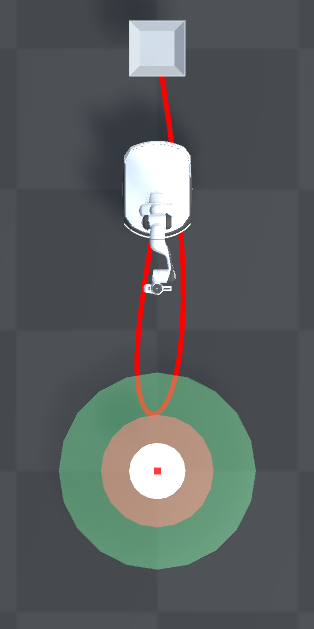}
\caption{Turn}
\label{fig:Turn}
\end{subfigure}
\begin{subfigure}{.415\linewidth}
\centering
\includegraphics[width=\linewidth]{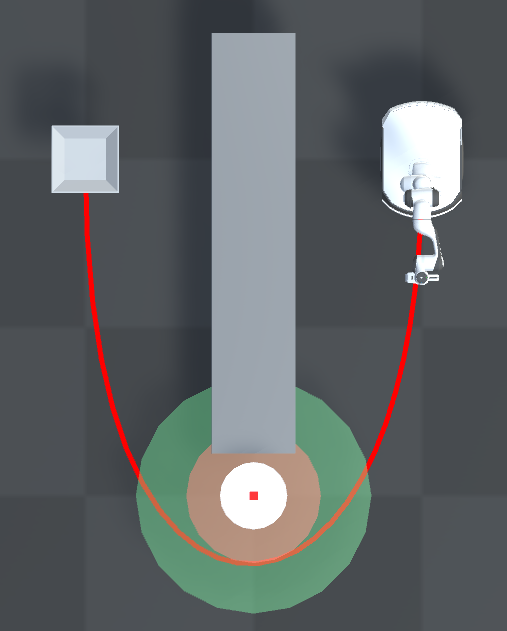}
\caption{Obstructed Turn}
\label{fig:ObstructedTurn}
\end{subfigure}
\caption{Three example pick-and-place scenarios. The green circles represent the area in which the robot can reach the object. The red circles indicate the area in which the robot will collide with the table the object is resting on. The red paths present example trajectories the robot could transit to perform the task.}
\label{fig:Scenarios}
\vspace{-15pt}
\end{figure}

We identify the three scenarios presented in Fig. \ref{fig:Scenarios} as representative of common pick-and-place tasks. These scenarios do not represent an exhaustive list, but illustrate how the arrangement of the environment impacts the amount of time the object will be within reach of the robot as it drives past. For the obstructed turn scenario, there is significant time for the robot to recover from a failed grasp while it drives around the object toward the drop point. By comparison, the turn environment presents a very limited manipulation window. To gracefully recover in all scenarios, the goal for the mobile base should be selected, and adjusted online, such that the object is kept within range until the grasp is completed. We focus only on minor manipulation failures where the grasp can be immediately reattempted. Larger failures, such as the object falling to the floor, require additional recovery behaviours.

\section{Related Works}
The base-placement problem for mobile manipulation determines a collision-free pose for the robot base from which to perform a manipulation task \cite{SerajiBasePlacement}. Approaches typically aim to generate solutions that are optimal against some other metric such as manipulability \cite{VahrenkampRobotPlacement, Jauhri}, or stiffness of the robot \cite{Fan}. Other approaches aim to minimise task time by calculating base poses from which multiple targets can be reached without repositioning the base between tasks \cite{Paus, Xu}. Time efficiency can be optimised on multi-step tasks by choosing a pose based on where the robot must go after the immediate target \cite{Reister, Harada}. Reactivity can be achieved by frequently recomputing the optimal base pose \cite{Reister}.

Although the system presented in \cite{Reister} selects optimal base placements considering both the immediate and next target in order to reduce overall execution time for pick-and-place tasks, it does not explore further performance improvements possible by performing the pick and place actions while the robot is still in transit. Instead, the base comes to a stop while the robot performs the manipulation. A benefit of this approach is that in the event of a grasp failure it is trivial for the arm to reset and reattempt the task. By comparison, methods that explore manipulation on-the-move including those presented in \cite{ThakarTimeOptimal, ThakarManipulatorMotionPlanning, ThakarUncertainty, Colombo, Zimmermann, XuOptimizationMotionPlanner} cannot easily reattempt a task in the event of a failure because the robot may be driving away and out of reach of the object. 

We wish to unify the benefits of both on-the-move and stationary grasping approaches. In particular, we desire a system that will perform tasks on-the-move where possible, but slow down or stop when a manipulation failure requires increased time near a target. 

\section{Approach}
\label{section:Approach}
We divide the base control system into two components: a base placement module and a base controller. 

\subsection{Base Control}
The base controller is a modified version of the Short Term Aborting A* (STAA*) reactive mobile robot controller presented in \cite{STAA}. This system plans global paths with an A* search through a visibility graph, and generates the optimal path to a given pose, avoiding obstacles in the scene. This global path is intersected with a local grid around the robot to provide an intermediate target for the base. The system generates a plan towards the intermediate target which is aborted either when the goal is reached, or a timer expires. By aborting the search and enforcing an update frequency the system achieves real-time reactive control. 

We have introduced several modifications to the controller presented in \cite{STAA} to improve its performance in a manipulation on-the-move scenario. 

\subsubsection{Goal Orientation} The most important addition to STAA* is the inclusion of  an orientation to the goal state. Where STAA* considers driving to a point only, we include orientation which enables poses to be achieved that smoothly connect the immediate target with the next goal. 

\begin{figure}[t]
\centerline{\includegraphics[width=\linewidth]{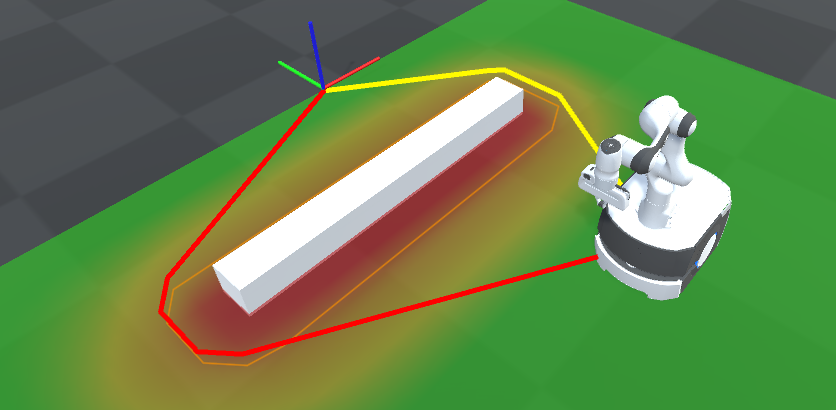}}
\caption{A comparison of global paths generated by our modified STAA* (path in red) and the original (shown in yellow) for an example goal pose. The red $x$ axis of the target frame represents the desired forward direction.}
\label{fig:GlobalPath}
\vspace{-15pt}
\end{figure}

\subsubsection{Rotation in Global Planner Cost} The addition of orientation to the goal state requires modification to the node cost computation used in the global A* search. STAA* uses only the cumulative distance between nodes along a path to compute the cost. Instead, we consider a PathRTR metric that estimates the time required to both translate and rotate between nodes along a path. Further explanation of the PathRTR metric is presented in \cite{STAA}. Fig. \ref{fig:GlobalPath} illustrates the value of including rotation costs in the global path planner. For the scenario shown, the path in red generated by our modified version encourages the robot to drive a smooth curve around the obstacle connecting the start and end pose. Without rotation costs the shortest path passes the obstacle on the other side and requires significantly more turning.

\subsubsection{Search Termination Conditions} The implementation of STAA* presented in \cite{STAA} terminates its search when an explored node is sufficiently close to the goal. To perform manipulation on-the-move, we want to encourage the robot to drive through the current target at high speed. Therefore, we also terminate when the path between a node and its parent passes sufficiently close to the goal.

\subsubsection{Reduction of Proximity Grid Penalty} STAA* includes a penalty on nodes based on their proximity to obstacles using an inflated occupancy grid. However, to complete mobile manipulation tasks such as picking and placing objects from a table, the robot must necessarily travel close to the table while interacting with objects on it. For example, in Fig. \ref{fig:BasePlacement}, the occupancy grid is represented by the colour of the ground around the robot, with green representing free space, and red representing occupied space. In this case, the target pose for the base from which it can perform the grasp is in a region with moderate proximity. The penalty applied to nodes near obstacles inhibits the exploration of states near the goal. To limit this effect, we reduce the weight of the proximity grid cost based on the expected time to the goal. We scale the grid penalty by $k = \max{(0.1, \min{(t_h/3, 1)})}$ where $t_h$ is the estimated time until the goal is achieved.

\subsection{Base Placement}
\begin{figure}[t]
\centerline{\includegraphics[width=\linewidth]{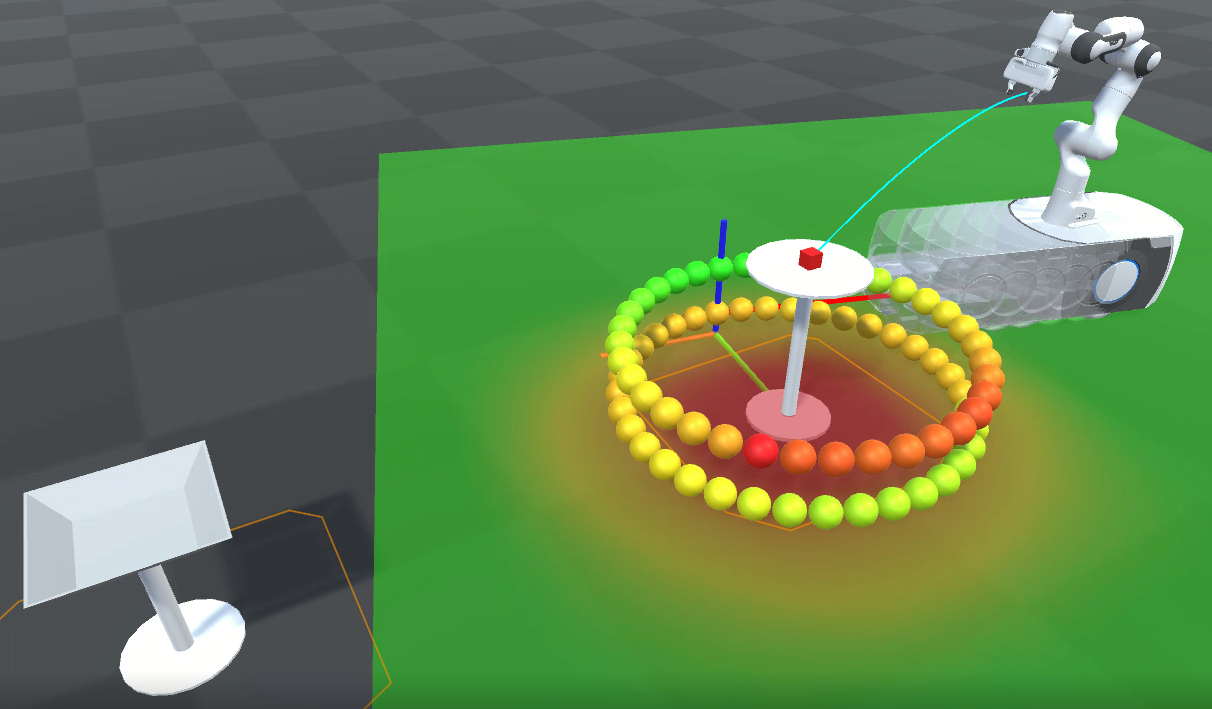}}
\caption{Illustration of sampled base placement results for a pick-and-place task. Each coloured sphere represents a candidate base placement around the target object. Those in the top ring represent a counter-clockwise heading, while those in the bottom face clockwise.}
\label{fig:BasePlacement}
\vspace{-15pt}
\end{figure}

The optimal base placement is selected from a discretised set by evaluating the path cost both from the current robot pose to candidate base placements as well as from the candidates to the drop point. Candidates are evenly spaced around the target object in 10\degree \ increments on a circle of radius 0.6 \si{m}, for a total of 36 possible base positions. Each position is assigned two possible orientations, the robot's forward vector can be tangential to the circle facing either clockwise or counter-clockwise which gives a total of 72 candidates. In Fig. \ref{fig:BasePlacement}, each coloured sphere represents a candidate, with those in the top ring possessing a counter-clockwise heading, and those in the bottom facing clockwise. 

The path cost for each candidate is evaluated using the global planner from the base control system. The highest-ranked pose from the candidate set is used as the goal for navigation. The result is a method that selects a candidate base placement that is within manipulation range of the target and efficiently connects the current robot pose with the drop pose. For example, the spheres in Fig. \ref{fig:BasePlacement} are coloured based on their path cost, with green representing the lowest cost. The selected base pose (shown by a target frame) encourages the robot to travel counter-clockwise around the object driving towards the drop point while it completes the grasp.

To enable failure recovery, the desired base placement should be continually updated in response to the current state of the robot. If more time is required to recover from a failed grasp, the system can then reevaluate the optimal base pose which is within manipulation range of the object, but as close as possible to the drop point. We evaluate each of the candidates on every controller step (20 \si{Hz}).

\section{Baselines}

We compare our method with two baselines that represent common approaches to mobile manipulation. All methods share the arm control method presented in \cite{MotM}.

\subsection{Conventional Reactive}
The first baseline is representative of several reactive controllers for mobile manipulation such as those presented in \cite{HavilandHolistic, Logothetis, Arora, Spahn}. In this baseline, the base placement method proposes the collision-free pose within manipulation range of the object
that is closest to the robot's current position. The base is controlled to this pose using the Timed Elastic Band implementation provided by the ROS Navigation stack\footnote{\url{https://wiki.ros.org/navigation}} which provides reactive control and obstacle avoidance. 

\subsection{Planned On-The-Move}
The planned baseline is similar to the proposed method but does not recompute the optimal base pose. Effectively, the base follows a precomputed path from the initial state to the drop point. This is representative of planning-based on-the-move approaches such as \cite{ShanMotionPlanning, ThakarTimeOptimal, ThakarManipulatorMotionPlanning, ThakarUncertainty, Colombo, Zimmermann, XuOptimizationMotionPlanner}. It should be noted that these approaches use a planned trajectory for the arm motion which does not allow for repeat grasp attempts.

\section{Experiments}

Experiments are conducted with the three arrangements of pick-and-place locations presented in Fig. \ref{fig:Scenarios}. Failures were simulated by artificially causing grasp attempts to fail for a set amount of time after the first grasp attempt is made. This allows for investigation of the expected performance when performing tasks of varying complexity, where it may take longer to recover from failures in more challenging tasks. We test each system in each scenario with grasp failure delays between 0 and 10 seconds, where a failure delay of 0 seconds represents success on the first attempt. For each trial we record the total task execution time from the start of the robot motion to the dropping of the object.

\section{Results}

\begin{figure}[t]
\centerline{\includegraphics[width=\linewidth]{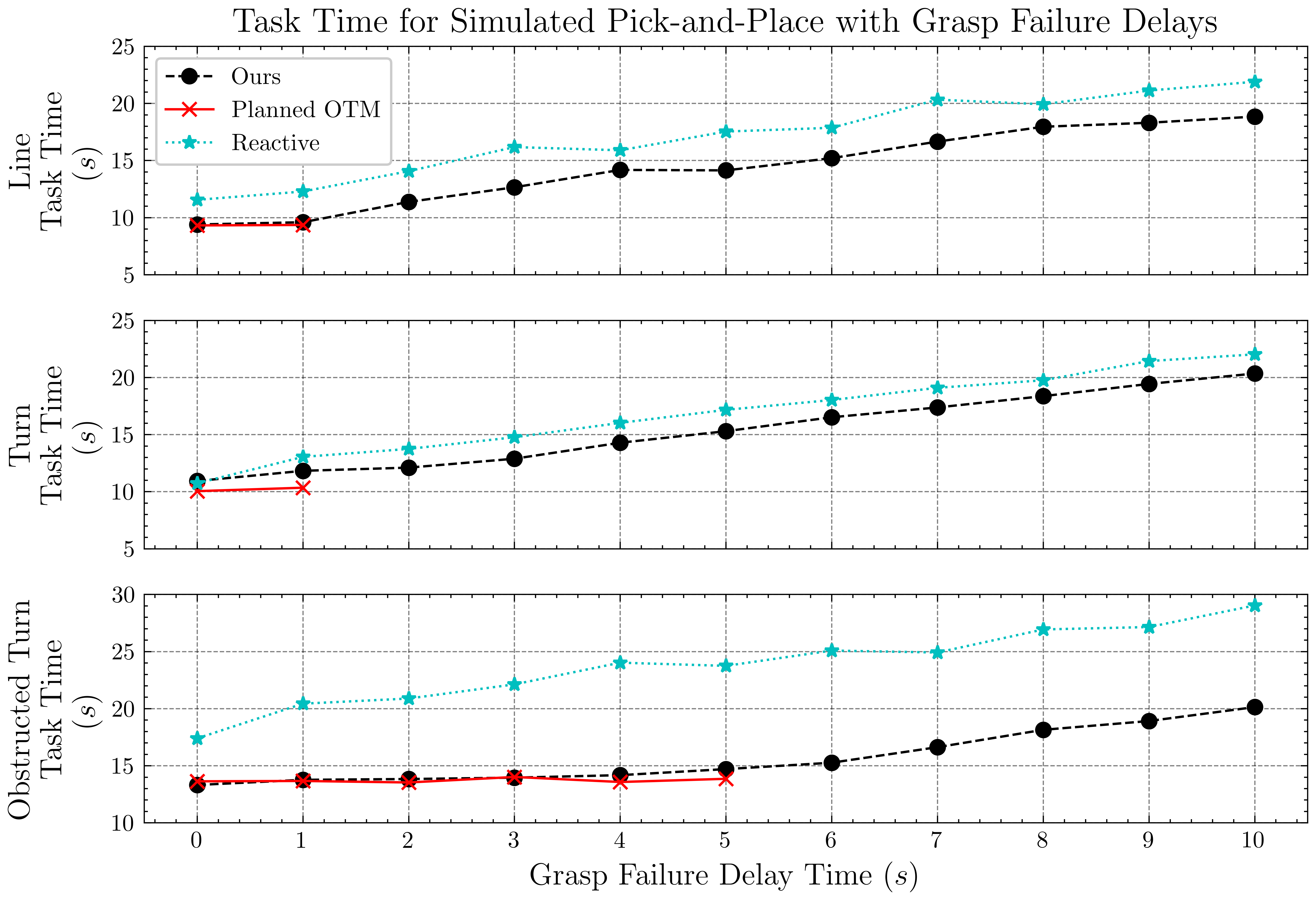}}
\caption{Results for simulated pick-and-place tasks with varying grasp delays in three different scenarios.}
\label{fig:Results}
\vspace{-15pt}
\end{figure}

Fig. \ref{fig:Results} presents the total task execution time for each of the methods in each scenario with varied delays caused by grasp failures. Results are only shown for successful task completions, and the lack of results for the planning baseline at increased failure delay times is indicative of the robot failing to grasp the object while driving past. 

In all cases, the on-the-move methods complete the task in less time than the reactive baseline. This is expected, however, it is interesting to note that the magnitude of the improvement is dependent on the task. For example, in the turn task (Fig. \ref{fig:Turn}), the base placements and subsequent paths taken by the reactive controller are relatively similar to those generated by our method because the optimal path involves driving straight to the target, turning around, and driving straight back. In this case, there is little opportunity for the on-the-move methods to make significant progress toward the next target while performing the manipulation. By contrast, in the task with an obstacle presented in Fig. \ref{fig:ObstructedTurn} the robot can drive a significant distance toward the drop point while the object remains within reach. 

The execution times for the reactive baseline increase linearly with increasing failure delay time. This is expected because the robot comes to a stop while it attempts to grasp the object and stays stationary until the grasp is completed. The robot can wait indefinitely providing the necessary time to recover from any failure, and consequently, it is successful in all trials. 

The planned on-the-move baseline performs similarly to our proposed method when the initially calculated optimal path stays within the manipulation range for longer than the failure delay time. However, for the tasks presented in Fig. \ref{fig:Line} and Fig. \ref{fig:Turn}, this is only the case up to a delay of 1 second. Longer failures result in the robot driving away from the object before the grasp has been completed. For the obstructed turn task (Fig \ref{fig:ObstructedTurn}) the path naturally orbits the object for around 5 seconds. Consequently, the planned baseline can successfully complete the task with failure delay times of up to 5 seconds. However, as the delay time increases further, the robot drives away before the task is completed. 

Fig. \ref{fig:VideoFrames} presents snapshots from videos of each system performing a pick-and-place task where the robot takes 6 seconds to recover from a grasp failure. At $t$ = 12\si{s}, the proposed method slows the robot to complete the grasp. The planned on-the-move baseline continues moving toward the drop point and is unable to recover from the grasp failure while still in range of the object, resulting in a failed task. The reactive baseline provides sufficient time near the object to recover from the failure, but the base does not make progress toward the drop point while the grasps are attempted, increasing execution time for the pick-and-place task.

The proposed method combines the benefits of both of the baselines. It performs tasks on-the-move with improved execution times when the robot can make progress toward the drop point while attempting to grasp the object. However, by updating the optimal base pose in real-time, the system compensates for delay times that exceed the typical transit time of driving past the object. Instead, the robot will slow down or come to a stop positioned as close to the drop point as possible, allowing time for the grasp controller to recover from failures, grasp the object, and complete the task.

\section{Conclusion}
Our proposed reactive base control system provides a practical solution for handling failures in mobile manipulation tasks performed on-the-move. However, many open questions remain in this space. In this work the robot has no knowledge of how long it might take to recover from a failure. Online prediction of this information could be used to further improve the system. Acceleration limits of the robot also have an important impact on the optimal solution to a given scenario. For example, a robot with a very limited acceleration might achieve improved performance on the presented scenarios by orbiting the object where possible instead of coming to a stop to complete the grasp. These questions present opportunities to further develop mobile manipulation systems that can operate efficiently in complex environments and adapt to uncertainties and failures.

\bibliographystyle{IEEEtran}
\bibliography{references}

\end{document}